\documentclass[
]{ceurart}
\pdfoutput=1
\usepackage{rotating}

\begin{document}

\copyrightyear{2021}
\copyrightclause{Copyright for this paper by its authors.
  Use permitted under Creative Commons License Attribution 4.0
  International (CC BY 4.0).}

\conference{DeFactify'23: SECOND WORKSHOP ON MULTIMODAL FACT CHECKING AND HATE SPEECH DETECTION,
  FEBRUARY, 2023, Washington, DC, USA}

\title{NUAA-QMUL-AIIT at Memotion 3: Multi-modal Fusion with Squeeze-and-Excitation for Internet Meme Emotion Analysis}

\author[1,3]{Xiaoyu Guo}[%
orcid=0000-0002-0901-2222,
email=xiaoyu.guo@nuaa.edu.cn,
]
\address[1]{College of Economics and Management, Nanjing University of Aeronautics and Astronautics (NUAA),
  Nanjing, Jiangsu, China}

\author[1]{Jing Ma}[%
orcid=0000-0001-8472-2518,
email=majing5525@126.com,
]

\author[2]{Arkaitz Zubiaga}[%
orcid=0000-0003-4583-3623,
email=a.zubiaga@qmul.ac.uk,
url=http://www.zubiaga.org/,
]
\address[2]{School of Electronic Engineering and Computer Science, Queen Mary University of London (QMUL), London, UK}
\address[3]{Advanced Institute of Information Technology (AIIT), Peking University, Hangzhou, Zhejiang, China}
\begin{abstract}
  This paper describes the participation of our NUAA-QMUL-AIIT team in the Memotion 3 shared task on meme emotion analysis. We propose a novel multi-modal fusion method, Squeeze-and-Excitation Fusion (SEFusion), and embed it into our system for emotion classification in memes. SEFusion is a simple fusion method that employs fully connected layers, reshaping, and matrix multiplication. SEFusion learns a weight for each modality and then applies it to its own modality feature. We evaluate the performance of our system on the three Memotion 3 sub-tasks. Among all participating systems in this Memotion 3 shared task, our system ranked first on task A, fifth on task B, and second on task C. Our proposed SEFusion provides the flexibility to fuse any features from different modalities. The source code for our method is published on \url{https://github.com/xxxxxxxxy/memotion3-SEFusion}.
\end{abstract}


\maketitle

\section{Introduction}

With the rapid increase in the amount of online information, automated processing of the content can help alleviate the otherwise burdensome task of sifting through all the information. One of the prevalent forms of online information is the one spread as internet memes. An internet meme is a concise and often humorous means of sharing information online, generally
communicated as an image with text embedded \cite{guo2021a}. In recent years, internet memes have become prevalent as a means to share opinions through different Internet platforms such as social media \cite{guo-etal-2020-nuaa}.

Generally, internet memes combine two modalities: image and text. While the content of memes can be useful and important to be processed through automated means, much of the existing research has limited to text, with less attention paid to the analysis of memes, as is the case in our work focused on meme emotion analysis. The key challenge of meme emotion analysis is achieving an effective combination of the text and image features extracted by pre-trained models. Existing fusion methods mainly use an attention mechanism to map the features of the different modalities (e.g. \cite{li2020attention,sun2021deep,chen2022research}). An important aspect to be considered when combining the modalities is determining the weight in each case, as it varies from case to case where either the text or the image plays a more significant role. One can then combine the modalities by multiplying and subsequently aggregating the inferred weights with their associated embeddings. With our work, the main objective is to optimise the learning of the weights of each modality through the use of neural network models.

Our work builds on an approach introduced by \citet{hu2018squeeze}, who proposed a squeeze-and-excitation block to learn the channel dependencies of an image, which can be applied to a variety of deep neural networks leading to improved classification performance. Inspired by this work, we consider utilizing squeeze-and-excitation to learn the modal dependencies of multi-modal data. The squeeze-and-excitation block cannot be applied directly to fuse features of different modalities and hence we adopt the framework in order to adapt it to our multi-modal fusion task.

In this article, we propose Squeeze-and-Excitation Fusion (SEFusion), a novel multi-modal fusion method, and apply it to fuse text features and image features extracted from internet memes. Through testing it on the Memotion 3 shared task, our SEFusion system achieved the top rank in task A of the competition, with an F1 score of 0.3441. SEFusion system also ranked second on task C.

The rest of this paper is organized as follows. In the next section, we describe the Memotion 3 task and prior work on emotion classification in memes. Then in Section 3, we propose SEFusion, a novel multi-modal fusion method, and embed it into our system to classify memes. We then employ the method to analyze memes in task A, task B, and task C in Section 4. In Section 5, we discuss the results of our experiments. Finally, we conclude with the findings of this research and suggest directions for future work.

\section{Background}

\subsection{The Memotion 3 Shared Task}

Memotion 3 \cite{mishra2023memotionoverview} is the 3rd edition of the series of Memotion shared tasks focused on meme emotion analysis. The previous editions-Memotion 1 \cite{sharma2020semeval} and Memotion 2 \cite{patwa2022findings} provided annotated datasets \cite{ramamoorthy2022memotion} and have brought attention to the analysis of memes. This edition consists of three subtasks: (i) Task A in classifying an internet meme according to its expressed emotion as positive, negative, or neutral, (ii) Task B in identifying whether an internet meme is sarcastic, humorous, offensive, or motivational as a multi-label classification task, and (iii) Task C in quantifying the scales of each type in task B.

\subsection{Related Work on Meme Emotion Analysis}

Previous research on meme emotion analysis mainly focuses on emotion classification, identifying the type of emotion expressed, and detecting hateful memes, which are all part of the Memotion 3 task. \citet{wu2018slangsd} focus on text memes and add slang and sentiment lexica as extra information to improve the performance of meme emotion classification. \citet{amalia2018meme} firstly use OCR Tesseract to extract text from image memes and then classify the extracted text into positive or negative employing the Naive Bayes classifier, which achieves a competitive accuracy of 75\%. As for identifying the type of emotion expressed, \citet{costa2015reality} propose to use a Maximum Entropy classifier to recognize humorous text memes. Their model achieved high performance for the negative class, with substantially lower performance for the positive class. \citet{sabat2019hate} use BERT and VGG-16 to process texts and images for hateful meme detection. They apply both early fusion and late fusion methods to combine text and image.

Most recently, \citet{nayak2022detection} employ various machine learning models to automatically detect hate in internet memes. \citet{ouaari2022multimodal} use neural networks to extract features of internet memes and train a classifier to identify the sentiment expressed in memes. \citet{fersini2022misogynous} posit that hateful content is expressed through memes and, to support with their detection, they utilize unimodal and multimodal approaches to identify misogynous memes.

In summary, previous research on meme emotion analysis has considered a wide range of traditional machine learning, more contemporary deep learning models, and well-established feature fusion methods. To our knowledge, existing research does not combine features of different modalities by learning the weight of different modalities automatically. By studying this combination in the context of memes, our study introduces a novel fusion method that attempts to learn the weights of different modalities.

\section{System Overview}

\subsection{Breaking Down the Task into Subtasks}

The shared task consists of three subtasks which, in our case, we envisaged as nine classification sub-tasks (one for task A, four for task B, and four for task C). This is because tasks B and C require making four predictions each, which we considered to tackle separately. Hence, we assign specific names to each of these classification sub-tasks (A, B1-4, C1-4), as shown in Table \ref{table1}. Throughout the experimentation period, we observed that there was no difference between B4 and C4, so we regard these two as the same sub-task, hence reducing it to eight sub-tasks.

\begin{table}[h]
\centering
\begin{tabular}{lll}
\hline
Task                    & Sub-task & Content                                                                         \\ \hline
task A                  & task A   & Classify a meme as positive, negative, or neutral.                              \\ \hline
\multirow{4}{*}{task B} & task B1  & Classify a meme as humorous or not.                                             \\
                        & task B2  & Classify a meme as sarcastic or not.                                            \\
                        & task B3  & Classify a meme as offensive or not.                                            \\
                        & task B4  & Classify a meme as motivational or not.                                         \\ \hline
\multirow{4}{*}{task C} & task C1  & Quantify a meme as not funny, funny, very funny, or hilarious.                  \\
                        & task C2  & Quantify a meme as not sarcastic, general, twisted meaning, or very twisted.    \\
                        & task C3  & Quantify a meme as not offensive, slight, very offensive, or hateful offensive. \\
                        & task C4  & Quantify a meme as motivational or not.                                         \\ \hline
\end{tabular}
\caption{Nine classification sub-tasks of Memotion 3. Note: task B4 and task C4 are regarded as the same sub-task, hence reducing it to eight.}
\label{table1}
\end{table}

\begin{figure}[h]
  \centering
  \includegraphics[width=\linewidth]{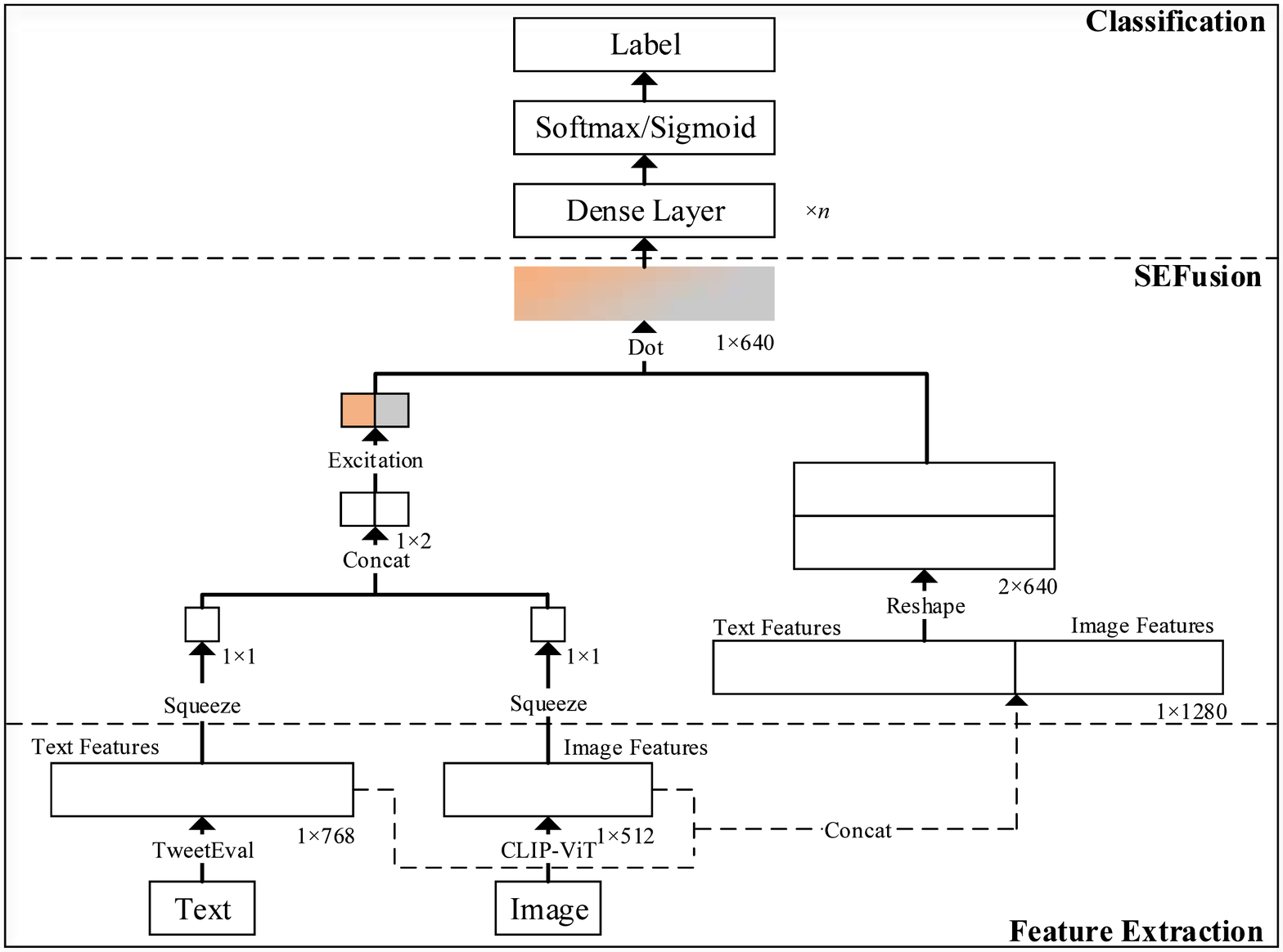}
  \caption{Multi-modal fusion with squeeze-and-excitation for internet meme classification. Firstly, TweetEval \cite{barbieri2020tweeteval} and CLIP-ViT \cite{zhai2019large} are used to extract text features and image features, respectively. Secondly, we use our proposed SEFusion to fuse the features of binary modalities. Finally, $n$ fully connected layers with the activation of \emph{sigmoid} (or \emph{softmax}) are utilized for classification.}
  \label{figure1}
\end{figure}

We therefore approach the task as eight classification sub-tasks (A, B1-4, C1-3). All sub-tasks use the same system framework, which is depicted in Figure \ref{figure1}. First, we extract text and image features. Subsequently, we fuse these features together with our proposed SEFusion. Finally, the fused features are sent to dense layers with a proper activation to produce the category label.

\subsection{Feature Extraction}
Internet memes are in a concise form\cite{hakokongas2020persuasion} and thus it is uneasy to extract enough features from themselves. The universal and semantic features could be learned from the large corpus during pre-training. Therefore, we choose pre-trained models to extract features from internet memes.

\subsubsection{Data Pre-processing}

Texts extracted from memes contain many user names, as strings that start with ``@” followed by other characters representing the user name. Given that this set of characters will likely not be meaningful for the meme emotion analysis, we replace them with a generic token ``@user''. In addition, several memes have watermarks, showing a link for their creator or origin. We replace these links with the generic token ``http”. For the image, we perform the default pre-processing of the pre-trained model\footnote{https://huggingface.co/laion/CLIP-ViT-B-32-laion2B-s34B-b79K/blob/main/preprocessor\_config.json}.

\subsubsection{Text Feature Extraction}

We use TweetEval \cite{barbieri2020tweeteval} to extract the text features\footnote{https://github.com/cardiffnlp/tweeteval/blob/main/TweetEval\_Tutorial.ipynb}. The pre-trained model we choose is cardiffnlp/twitter-roberta-base. We take the average of the extracted features as the representation of each item of meme text. The text features are denoted as $\mathbf{X}_{\mathrm{t}}$, $\mathbf{X}_{\mathrm{t}} \in \mathrm{R}^{1^{\times} 768}$.

\subsubsection{Image Feature Extraction}

We use CLIP-ViT \cite{zhai2019large} to extract the image features. The pre-trained model we use is laion/CLIP-ViT-B-32-laion2B-s34B-b79K. We then perform L2 normalization to get the final image features $\mathbf{X}_{\mathrm{i}}$, $\mathbf{X}_{\mathrm{i}} \in \mathrm{R}^{1^{\times} 512}$.

\subsection{Squeeze-and-excitation Fusion}

SEFusion is a computational unit that can be built upon multi-modal features. Since the output of the multi-modal model is produced by a summation through all modalities, modal dependencies are significant for multi-modal feature fusion. Our proposed fusion method can learn the relationships between modalities and explicitly model modal interdependencies. The procedure of SEFusion is shown in the middle of Figure \ref{figure1}.

We next describe the two components of SEFusion, squeeze and excitation.

\subsubsection{Squeeze}

In order to tackle the issue of exploiting modal dependencies, we consider learning the weight of each modality. We first perform dimension reduction on the text and image features. We use the dense layer and set the unit as 1 to linearly squeeze text features into a vector $z_{mt}$, $z_{mt} \in \mathrm{R}^{1^{\times} 1}$, which is different from the squeeze procedure in \cite{hu2018squeeze} since the feature dimension in our case is different from the dimension produced by the convolutional operator in \cite{hu2018squeeze}. We also get $z_{mi}$ with the same operation on image features. Next, we concatenate $z_{mt}$ with $z_{mi}$ and get $z$, $z \in \mathrm{R}^{1^{\times} 2}$. The procedure is shown as:

\begin{equation}
 z_{m t}=\mathbf{F}_{s q}\left(\mathbf{X}_t, \mathbf{W}_1\right)=\mathbf{W}_1 \mathbf{X}_t,
\end{equation}

\begin{equation}
 z_{m i}=\mathbf{F}_{s q}\left(\mathbf{X}_i, \mathbf{W}_2\right)=\mathbf{W}_2 \mathbf{X}_i,
\end{equation}

\begin{equation}
 \mathbf{z}=\operatorname{Concat}\left(z_{m t}, z_{m i}\right).
\end{equation}

\subsubsection{Excitation}

To make use of the information aggregated in the squeeze operation, we follow it with a second operation that aims to fully capture modal-wise dependencies. Following \citet{hu2018squeeze}, we opt to employ a simple gating mechanism with a sigmoid activation:

\begin{equation}
 \mathbf{s}=\mathbf{F}_{e x}(\mathbf{z}, \mathbf{W})=\sigma(g(\mathbf{z}, \mathbf{W}))=\sigma\left(\mathbf{W}_4 \delta\left(\mathbf{W}_3 \mathbf{z}\right)\right),
\end{equation}
where $\delta$ refers to the ReLU \cite{nair2010rectified} function, $\mathbf{W}_{\mathrm{3}} \in \mathrm{R}^{2^{\times} 1}$, $\mathbf{W}_{\mathrm{4}} \in \mathrm{R}^{1^{\times} 2}$, and $\mathbf{s}$ is the learned vector of weights for the modalities.

Next, we apply the weight vector to the multi-modal features for computing the fused features. Considering that the dimensionality of the text and image features are different, we concatenate these features and directly reshape the concatenated features into $\mathbf{X}^{\prime}\left(\mathbf{X}^{\prime} \in \mathrm{R}^{2 \times 640}\right)$ in order to apply the operation matrix multiplication on $\mathbf{s}$ and $\mathbf{X}^{\prime}$ easily. The final fused feature is calculated by:

\begin{equation}
 \mathbf{X}=\operatorname{Concat}\left(\mathbf{X}_{\mathrm{t}}, \mathbf{X}_{\mathrm{i}}\right),
\end{equation}

\begin{equation}
 \mathbf{X}^{\prime}=\operatorname{Reshape}(\mathbf{X}),
\end{equation}

\begin{equation}
 \mathbf{X}_{\text {fusion }}=\mathbf{s} \mathbf{X}^{\prime},
\end{equation}
where $\mathbf{X} \in \mathrm{R}^{1^{\times} 1280}$, $\mathbf{X}^{\prime} \in \mathrm{R}^{2^{\times} 640}$, and $\mathbf{X}_{\text {fusion}} \in \mathrm{R}^{1^{\times} 640}$.

\emph{Discussion}. After reshaping, we got $\mathbf{X}^{\prime}$, whose first row contains partial features of the text while the second row contains the combination of image features and the remaining text features. When applying the operation matrix multiplication on $\mathbf{s}$ and $\mathbf{X}^{\prime}$, we put the image weight on partial features of the text, which may bring some dispute. It is also acceptable to unify the feature dimension of each modality by following a dense layer.
\subsection{Classification}

The fused layer is used as the input to $n$ fully connected layers, where $n$ is a hyper-parameter and needs to be adjusted for different sub-tasks. The fully connected layers are followed by the activation of \emph{sigmoid} (or \emph{softmax}) for generating the probability of the image pertaining to a class.

\section{Experimental Setup}
\subsection{Dataset}

The dataset used for our experiments was released by the organizers of the Memotion 3 task \cite{mishra2023memotion3}. Each entry in the dataset contains the following fields: image, text, and label. The field of label varies for the different tasks.
The dataset contains a total of 10,000 samples, including 7,000 for training, 1,500 for validation, and 1,500 for test. For the experimentation, we rely on the training, validation, and test data as split by the organizers. Tables \ref{table2}-\ref{table4} show the distribution of labels of different tasks across training, validation, and test sets.

\begin{table}[h]
\centering
\begin{tabular}{lllll}
\hline
         & Train       & Validation & Test      & Sum         \\ \hline
Positive & 2,275(33\%) & 341(23\%)  & 586(39\%) & 3,202(32\%) \\
Neutral  & 2,970(42\%) & 579(39\%)  & 533(36\%) & 4,082(40\%) \\
Negative & 1,755(25\%) & 580(39\%)  & 381(25\%) & 2,716(27\%) \\
Sum      & 7,000       & 1,500      & 1,500     & 10,000       \\ \hline
\end{tabular}
\caption{The distribution of task A labels.}
\label{table2}
\end{table}

\begin{table}[h]
\centering
\begin{tabular}{llll|lll}
\hline
    & \multicolumn{3}{l|}{Humor (task B1)}     & \multicolumn{3}{l}{Sarcastic (task B2)}    \\
    & Train        & Validation  & Test        & Train        & Validation   & Test         \\ \hline
Yes & 5,990(86\%)  & 1,401(93\%) & 1,389(93\%) & 5,524(79\%)  & 1,377(92\%)  & 1,367(91\%)  \\
No  & 1,010(14\%)  & 99(7\%)     & 111(7\%)    & 1,476(21\%)  & 123(7\%)     & 133(9\%)     \\
Sum & 7,000        & 1,500       & 1,500       & 7,000        & 1,500        & 1,500        \\ \hline
    & \multicolumn{3}{l|}{Offensive (task B3)} & \multicolumn{3}{l}{Motivational (task B4)} \\
    & Train        & Validation  & Test        & Train        & Validation   & Test         \\ \hline
Yes & 2,736(39\%)  & 859(57\%)   & 825(55\%)   & 830(12\%)    & 43(3\%)      & 56(4\%)      \\
No  & 4,264(61\%)  & 641(43\%)   & 675(45\%)   & 6,170(88\%)  & 1,457(97\%)  & 1,444(96\%)  \\
Sum & 7,000        & 1,500       & 1,500       & 7,000        & 1,500        & 1,500        \\ \hline
\end{tabular}
\caption{The distribution of task B labels.}
\label{table3}
\end{table}

\begin{table}[h]
\centering
\begin{tabular}{llll|lll|lll}
\hline
         & \multicolumn{3}{l|}{Scale of humor (task C1)}                                                                                                                        & \multicolumn{3}{l|}{Scale of sarcastic (task C2)}                                                                                                                      & \multicolumn{3}{l}{Scale of offensive (task C3)}                                                                                                                     \\
         & Train                                                  & Validation                                           & Test                                                 & Train                                                  & Validation                                           & Test                                                   & Train                                                  & Validation                                           & Test                                                 \\ \hline
Not      & \begin{tabular}[c]{@{}l@{}}1,010\\ (14\%)\end{tabular} & \begin{tabular}[c]{@{}l@{}}99\\ (7\%)\end{tabular}   & \begin{tabular}[c]{@{}l@{}}111\\ (7\%)\end{tabular}  & \begin{tabular}[c]{@{}l@{}}1,476\\ (21\%)\end{tabular} & \begin{tabular}[c]{@{}l@{}}123\\ (8\%)\end{tabular}  & \begin{tabular}[c]{@{}l@{}}133\\ (9\%)\end{tabular}    & \begin{tabular}[c]{@{}l@{}}4,264\\ (61\%)\end{tabular} & \begin{tabular}[c]{@{}l@{}}641\\ (43\%)\end{tabular} & \begin{tabular}[c]{@{}l@{}}675\\ (45\%)\end{tabular} \\
Slightly & \begin{tabular}[c]{@{}l@{}}3,393\\ (48\%)\end{tabular} & \begin{tabular}[c]{@{}l@{}}973\\ (65\%)\end{tabular} & \begin{tabular}[c]{@{}l@{}}928\\ (62\%)\end{tabular} & \begin{tabular}[c]{@{}l@{}}1,953\\ (28\%)\end{tabular} & \begin{tabular}[c]{@{}l@{}}977\\ (65\%)\end{tabular} & \begin{tabular}[c]{@{}l@{}}936\\ (62\%)\end{tabular}                                                      & \begin{tabular}[c]{@{}l@{}}1,935\\ (28\%)\end{tabular} & \begin{tabular}[c]{@{}l@{}}804\\ (54\%)\end{tabular} & \begin{tabular}[c]{@{}l@{}}762\\ (51\%)\end{tabular} \\
Mildly   & \begin{tabular}[c]{@{}l@{}}2,038\\ (29\%)\end{tabular} & \begin{tabular}[c]{@{}l@{}}375\\ (25\%)\end{tabular} & \begin{tabular}[c]{@{}l@{}}406\\ (27\%)\end{tabular} & \begin{tabular}[c]{@{}l@{}}3,021\\ (43\%)\end{tabular} & \begin{tabular}[c]{@{}l@{}}376\\ (25\%)\end{tabular} & \begin{tabular}[c]{@{}l@{}}403\\ (27\%)\end{tabular}                                                      & \begin{tabular}[c]{@{}l@{}}610\\ (9\%)\end{tabular}    & \begin{tabular}[c]{@{}l@{}}44\\ (3\%)\end{tabular}   & \begin{tabular}[c]{@{}l@{}}50\\ (3\%)\end{tabular}   \\
Very     & \begin{tabular}[c]{@{}l@{}}559\\ (8\%)\end{tabular}    & \begin{tabular}[c]{@{}l@{}}53\\ (4\%)\end{tabular}   & \begin{tabular}[c]{@{}l@{}}55\\ (4\%)\end{tabular}   & \begin{tabular}[c]{@{}l@{}}550\\ (8\%)\end{tabular}                                               & \begin{tabular}[c]{@{}l@{}}24\\ (2\%)\end{tabular}   & \begin{tabular}[c]{@{}l@{}}28\\ (2\%)\end{tabular} & \begin{tabular}[c]{@{}l@{}}191\\ (3\%)\end{tabular}    & \begin{tabular}[c]{@{}l@{}}11\\ (1\%)\end{tabular}   & \begin{tabular}[c]{@{}l@{}}13\\ (1\%)\end{tabular}   \\
Sum      & 7,000                                                  & 1,500                                                & 1,500                                                & 7,000                                                  & 1,500                                                & 1,500                                                  & 7,000                                                  & 1,500                                                & 1,500                                                \\ \hline
\end{tabular}
\caption{The distribution of task C labels.}
\label{table4}
\end{table}

\subsection{Parameter Setting}

Since the dataset is imbalanced, we employ the strategy of Logit Adjustment \cite{menon2020long} to overcome this problem. This strategy is implemented by changing the loss function and can be directly used in Keras.\footnote{https://kexue.fm/archives/7615} Therefore, we use sparse\_categorical\_crossentropy\_with\_prior as the loss function in our experiments. In addition, it is necessary to add the prior distribution of labels to the loss function. As the label distributions for the validation and test sets are not known during training, we use the label distribution of training sets. From Table \ref{table2}, we see that the labels of task A are distributed into 2,275 positive (33\%), 2,970 neutral (42\%), and 1,454 negative (25\%) instances. The distribution of other sub-tasks can be drawn from Tables \ref{table3} and \ref{table4}.

The batch size is set to 256, the learning rate is set to $1e^{-4}$, and Adam is used as the optimizer. For task A and task B, we use 2 dense layers; while for task C, we use 5 dense layers. We monitor the sparse categorical accuracy on the validation set to save the best model.

\subsection{Implementation}

We utilize Keras\footnote{https://keras.io/zh/}, the python deep learning library, to build the whole model structure. TweetEval \cite{barbieri2020tweeteval} and CLIP-ViT \cite{zhai2019large} are used to acquire the text and image representations through API provided by huggingface\footnote{https://huggingface.co/}.

\subsection{Evaluation Metrics}

We use weighted-F1 as the evaluation metric, which is the official metric proposed by the organizers. The weighted-F1 score is calculated by taking the mean of all per-class F1 scores while considering each class’s support, which is shown as:

\begin{equation}
 \mathrm{F} 1=\frac{2 \times \mathrm{P} \times \mathrm{R}}{\mathrm{P}+\mathrm{R}},
\end{equation}

\begin{equation}
 \text { weighted-F1 }=\sum_{i=1}^C p_i \mathrm{Fl}_i \text,
\end{equation}
where P and R stand for precision and recall, respectively. $p$ denotes the support proportion. $C$ is the total number of classes.

For task A, weighted-F1 can be used to evaluate directly. For task B and task C, we calculate the weighted-F1 score for each of the sub-tasks and then take an average of those scores to obtain an average-weighted-F1 score.

\section{Results}

Among all participating systems in this Memotion 3 task, our model achieved the 1st score on the evaluation of task A and the 2nd score on the evaluation of task C. The weighted-F1 scores and average-weighted-F1 scores for our proposed SEFusion are shown in Table \ref{table5}.

\begin{table}[h]
\centering
\begin{tabular}{llllllll}
\hline
                        &          & \multicolumn{3}{l}{Weighted-F1 scores} & \multicolumn{3}{l}{Average-weighted-F1 scores}                              \\
Task                    & Sub-task & Train      & Validation    & Test      & Train                   & Validation              & Test                    \\ \hline
Task A                  & Task A   & 0.3359     & 0.3643        & 0.3441    & -                       & -                       & -                       \\ \hline
\multirow{4}{*}{Task B} & Task B1  & 0.7598     & 0.8448        & 0.8344    & \multirow{4}{*}{0.6898} & \multirow{4}{*}{0.7885} & \multirow{4}{*}{0.7802} \\
                        & Task B2  & 0.7107     & 0.8283        & 0.8243    &                         &                         &                         \\
                        & Task B3  & 0.4621     & 0.5241        & 0.5177    &                         &                         &                         \\
                        & Task B4  & 0.8264     & 0.9569        & 0.9444    &                         &                         &                         \\ \hline
\multirow{4}{*}{Task C} & Task C1  & 0.4889     & 0.4765        & 0.4634    & \multirow{4}{*}{0.5490}  & \multirow{4}{*}{0.5917} & \multirow{4}{*}{0.5706} \\
                        & Task C2  & 0.3223     & 0.4707        & 0.4429    &                         &                         &                         \\
                        & Task C3  & 0.5584     & 0.4625        & 0.4317    &                         &                         &                         \\
                        & Task C4  & 0.8264     & 0.9569        & 0.9444    &                         &                         &                         \\ \hline
\end{tabular}
\caption{Weighted-F1 scores and average-weighted-F1 scores for SEFusion. Note: The weighted-F1 score of task B4 on the test set is different from our submission answer. We intended to figure out why the score of task B is as not good as task A and task C. We checked our code and found that we loaded the saved model of task B3 when we predicted the result of task B4. Therefore, we changed to the correct model to predict for task B4.}
\label{table5}
\end{table}

In Table \ref{table5}, we can conclude that our models are under-fitting except task C1 and task C3 since the evaluation scores on the training set are lower than on the validation set. To our best knowledge, under-fitting hardly happens on all the memotion datasets even using simple machine learning models. Under-fitting indicates that the performance of our model could be improved by training longer or adding extra layers to the network. The results of task C3 show that the model is over-fitting and we should cut the layers. For task C1, although there is a little overfitting, the extent of overfitting is acceptable.


In Table \ref{table5}, we also see that the performance of task B3 is lower than other sub-tasks in task B. All sub-tasks in task B are binary classification and they should be easier than task A and task C. However, the weighted-F1 score of task B3 is near to 0.5, which is the general baseline of binary classification. Given that the class proportion of task B3 varies less, we conclude that identifying the offensive memes is very hard using our existing features, and likewise classifying memes as positive, neutral, or negative.

\section{Conclusion}

In this paper, we propose SEFusion, a novel multi-modal fusion method to combine text and image features jointly for emotion classification in internet memes. Our method ranks first on task A and second on task C in Memotion 3 task.

Given the features extracted from memes, our proposed SEFusion applies squeeze and excitation, which are simple operations merely using fully connected layers with proper activations, reshaping, and matrix multiplication, to fuse text and image features. Like the Squeeze-and Excitation Block \cite{hu2018squeeze}, our proposed SEFusion is flexible and can be used to fuse other sets of features extracted through other models. In addition, SEFusion can fuse more than two types of features as long as the dimension is reshaped correctly.

Our work has some limitations and opens up avenues for future research. First, our model learned the weight vector for each modality, but the weight did not apply to the corresponding modality since we mixed the text and image features when reshaping the concatenated feature vector. We will consider unifying the feature dimension of different modalities before performing SEFusion. Second, internet meme emotion analysis is still in its infancy. Although our model ranks first in task A, its performance only slightly above the baseline model has room for improvement, which calls for more research, ideally jointly working with the adjacent tasks of detecting sentiment and hateful content from memes.

\section{Acknowledgments}

This study was supported by the National Natural Science Foundation of China under grant number 72174086. Xiaoyu Guo conducted this work while doing an internship at Advanced Institute of Information Technology, Peking University.


\bibliography{sample-ceur.bib}

\end{document}